\documentclass[sigconf]{acmart}

\AtBeginDocument{%
  }

\setcopyright{acmcopyright}
\copyrightyear{2023} 
\acmYear{2023} 
\setcopyright{acmlicensed}\acmConference[MM '23]{Proceedings of the 31st
ACM International Conference on Multimedia}{October 29-November 3,
2023}{Ottawa, ON, Canada}
\acmBooktitle{Proceedings of the 31st ACM International Conference on
Multimedia (MM '23), October 29-November 3, 2023, Ottawa, ON, Canada}
\acmPrice{15.00}
\acmDOI{10.1145/3581783.3611904}
\acmISBN{979-8-4007-0108-5/23/10}

\usepackage{multirow}
\usepackage{subfigure}

\begin{document}

\title{Uniformly Distributed Category Prototype-Guided Vision-Language Framework for Long-Tail Recognition}

\author{Siming Fu}
\authornote{Both authors contributed equally to this research.}
\email{fusiming@zju.edu.cn}
\author{Xiaoxuan He}
\authornotemark[1]
\email{Xiaoxuan_He@zju.edu.cn}
\affiliation{%
  \institution{Alibaba group}
  \city{Hangzhou Shi}
  \country{China}
}

\author{Xinpeng Ding}
\affiliation{%
  \institution{Hong Kong University of Science and Technology}
  \city{Hong Kong}
  \country{China}}
\email{xdingaf@connect.ust.hk}

\author{Yuchen Cao}
\affiliation{%
  \institution{Zhejiang University}
  \city{Hangzhou Shi}
  \country{China}
}
\email{Yuchen_Cao@zju.edu.cn}

\author{Hualiang Wang}
\authornote{Corresponding author.}
\affiliation{%
 \institution{Hong Kong University of Science and Technology}
 \city{Hong Kong}
 \country{China}}
\email{hwangfd@connect.ust.hk}

\renewcommand{\shortauthors}{Fu and He et al.}

\begin{abstract} 

Recently, large-scale pre-trained vision-language models have presented benefits for alleviating class imbalance in long-tailed recognition. However, the long-tailed data distribution can corrupt the representation space, where the distance between head and tail categories is much larger than the distance between two tail categories. This uneven feature space distribution causes the model to exhibit unclear and inseparable decision boundaries on the uniformly distributed test set, which lowers its performance.  To address these challenges, we propose the uniformly category prototype-guided vision-language framework to effectively mitigate feature space bias caused by data imbalance. Especially, we generate a set of category prototypes uniformly distributed on a hypersphere. Category prototype-guided mechanism for image-text matching makes the features of different classes converge to these distinct and uniformly distributed category prototypes, which maintain a uniform distribution in the feature space, and improve class boundaries. Additionally, our proposed irrelevant text filtering and attribute enhancement module allows the model to ignore irrelevant noisy text and focus more on key attribute information, thereby enhancing the robustness of our framework. In the image recognition fine-tuning stage, to address the positive bias problem of the learnable classifier, we design the class feature prototype-guided  classifier, which compensates for the performance of tail classes while maintaining the performance of head classes. Our method outperforms previous vision-language methods for long-tailed learning work by a large margin and achieves state-of-the-art performance.

\end{abstract}




\maketitle

\section{Introduction}
Real-world data often exhibits a long-tailed distribution, where a few head classes contain the majority of the data, while most tail classes have very few samples. This phenomenon is unfavorable for the practical application of deep learning models. As a result, many works have attempted to alleviate class imbalance from different perspectives, such as re-sampling training data~\cite{2019Decoupling, mahajan2018exploring, Fu_2022_ACCV}, re-weighting loss functions~\cite{elkan2001foundations, zhou2005training, sun2007cost, wang2022towards}, or using transfer learning methods~\cite{wang2017learning, chu2020feature, rsg}. Although these works have made significant contributions, they are still limited by relying solely on image patterns to solve the problem, ignoring intermodal relationships.

Undeniably, there are inherent connections between images and text descriptions of the same category, especially when it comes to visual concepts and attributes. Unlike image patterns that typically present specific low-level features of objects or scenes such as shape, color, and texture, text patterns can be summarized by experts as prior knowledge, which undoubtedly provides great help for model training when there are not enough images to learn a general classification representation for recognition. The recent advancements in vision-language models have resulted in the development of efficacious methodologies for acquiring robust representations that interconnect image and text modalities for long-tailed recognition. VL-TLR~\cite{tian2022vl} and BALLAD~\cite{ma2021simple} employ the multimodal potential of vision-language frameworks to devise powerful techniques that mitigate class imbalance.

\begin{figure}[!t]
\centering
\subfigure{
\centering \includegraphics[width=0.48\linewidth]{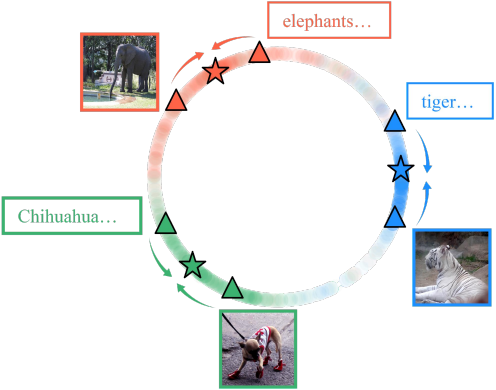}}
\subfigure{
\centering \includegraphics[width=0.48\linewidth]{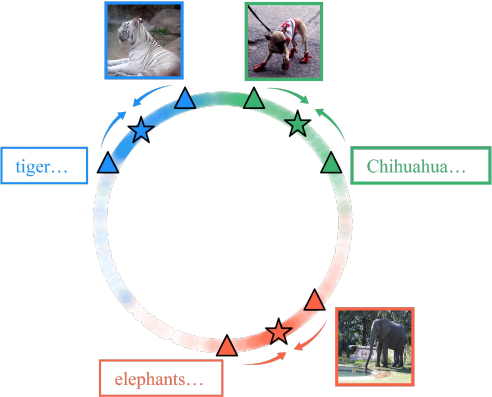}}
    \caption{A toy feature distribution comparison (left) before and (right) after implementing the uniformly distributed category prototype-guided image-text matching for three classes within ImageNet-LT. The elephant represents the head category, while the other two illustrate the tail categories. Prior to optimization, the representation space is dominated by the head category, leading to poor uniformity in learned category prototypes. Our approach improves performance by enhancing uniformity. Markers: Circles indicate representation space; triangles are text and image features; and pentagrams are category prototypes. The qualitative experiment visualization is shown in Appendix.~\ref{appendixB}.
    }
    \label{figure1}
\end{figure} 

The aforementioned studies~\cite{tian2022vl,ma2021simple} provide valuable perspectives on the capabilities of vision-language models in augmenting recognition performance for long-tailed datasets. However, these studies overlook a critical phenomenon: when the data distribution is imbalanced, the model's training process tends to emphasize uniformity loss between head and tail categories, while neglecting the uniformity loss between two tail categories~\cite{li2022targeted}. As illustrated in the left panel of Fig.~\ref{figure1}, the representation space is predominantly occupied by the head category (elephant), resulting in a substantially larger distance between the head and tail categories compared to the distance between the two tail categories. The greater the imbalance of long-tailed data, the greater the bias in the feature space distribution and the more uneven the distribution becomes. This uneven feature space distribution causes the model to exhibit unclear and inseparable decision boundaries on the uniformly distributed test set, which lowers its performance. This is the main reason why vision-language models applied to long-tailed recognition fail to achieve satisfactory performance. Furthermore, we notice that the text contains a significant amount of noise information, the representation learned by the model may incorporate these irrelevant details, leading to a decline in the quality and semantic relevance of the representation. This can result in inaccuracies and ambiguities in the inference process, thus leading to a misleading model. Last but not least, in the image recognition process, due to major classes having significantly more examples, classifiers are induced to focus on them to optimize evaluation metrics and accuracy. This leads to poor performance on tail classes, named the forward bias of the classifier. 

To address the aforementioned challenges, we propose a uniformly distributed category prototype-guided vision-language model framework, which effectively avoids the collapse of feature space representation quality and model performance caused by scarce data in tail categories. Specifically, we leverage the  “anchor text” to obtain uniformly distributed initialization category feature prototypes in the hypersphere feature space, and use these pre-generated prototypes as category targets. Furthermore, we design a visual-language representation framework based on category feature prototype contrastive learning, which is different from the conventional approach of directly aligning text and image features in contrastive learning. Instead, we match the cross-modalities features of corresponding categories to their target category prototypes during the stage of image-text matching. This ensures that the distribution of image and text features in the feature space is uniform and avoids being dominated and biased by the head categories. As depicted in the right panel of Fig.~\ref{figure1}, the acquired category prototypes demonstrate a uniform distribution, which consequently results in an enhancement of the model's performance.   In addition, based on the learned category feature prototypes, we design an extraneous text filter and  attribute enhancement module that enables the model to focus on key textual attributes while avoiding the learning of trivial correlations between image and text pairs.  In the final stage of image recognition fine-tuning, we design a class feature prototype-guided classifier to alleviate the positive bias issue inherent in conventional learnable classifiers~\cite{2019Decoupling}. The contributions of this paper can be summarized as follows:

\begin{itemize}
\item  We propose a uniformly distributed category prototype-guided visual-linguistic framework, in which the category prototype-guided mechanism for image-text matching guarantees a uniform distribution of image and text features in the feature space, preventing domination and bias by head categories.
\item We propose the irrelevant text filter and an attribute enhancement module that allows the model to focus on key textual attributes while avoiding the learning of trivial correlations between image and text pairs.
\item We design the category prototype-guided recognition classifier to alleviate positive bias in the inference process of head categories.
\item Our proposed method achieves state-of-the-art performance on long-tailed distributed datasets, including ImageNet-LT, Places-LT, and iNaturalist 2018.
\end{itemize}

\section{The proposed method}
\section{Related Work}
\textbf{Long-tailed Recognition.} Real-world data usually has an unbalanced distribution. To address the long-tailed class imbalance, massive deep long-tailed learning studies have been conducted in recent years, including class re-balancing, information augmentation, and module improvement. Class re-balancing seeks to re-balance the negative influence brought by the class imbalance in training sample numbers. Seesaw loss~\cite{wang2021seesaw} re-balances positive and negative gradients for each class with two re-weighting factors, i.e., mitigation and compensation. Label-Distribution-Aware Margin (LDAM)~\cite{cao2019learning} enforces class-dependent margin factors for different classes based on their training label frequencies, which encourages tail classes to have larger margins. The vMF classifier~\cite{wang2022towards} optimizes the class-wise features via vMF distributions to obtain uniform and aligned representations. Information augmentation seeks to introduce additional information
into model training so that the model performance can be improved for long-tailed learning. Routing Diverse Experts (RIDE)~\cite{wang2020long} introduced a knowledge distillation method to reduce the parameters of the multi-expert model by learning a student network with fewer experts. MiSLAS~\cite{zhong2021improving} proposed to use data mixup to enhance representation learning in the decoupled scheme. Module improvement improves network modules in long-tailed learning.  DRO-LT~\cite{samuel2021distributional} extended the prototypical contrastive learning with distribution robust optimization~\cite{goh2010distributionally}, which makes the learned model more robust to distribution shift. Notably, the aforementioned methods predominantly address long-tailed recognition by considering only image modalities, while seldom incorporating text modalities into the problem.

\noindent \textbf{Vision-Language Model.} Vision-language pre-training (VLP) is typically directed by specific vision-language objectives~\cite{radford2021learning, yao2021filip, li2022exploring, li2023clip}, facilitating the learning of image-text correspondences from extensive image-text pair datasets~\cite{schuhmann2022laion, schuhmann2021laion}. CLIP~\cite{radford2021learning} adopts an image-text contrastive objective, which learns by attracting paired images and texts while repelling unrelated ones in the embedding space. SimVLM~\cite{wang2021simvlm} considerably streamlines VLP by exclusively leveraging language modeling objectives on weakly aligned image-text pairs. ALIGN~\cite{jia2021scaling} introduces a straightforward dual-encoder architecture that learns to align visual and language representations from noisy image and text data, which can subsequently be applied to vision-only or vision-language task transfers. Furthermore, BLIP~\cite{li2022blip} presents a novel VLP framework that flexibly adapts to both vision-language understanding and generation tasks.

There have been several vision-language approaches designed for long-tailed recognition. BALLAD~\cite{ma2021simple} first continues pre-training the vision-language backbone through contrastive learning on a specific long-tailed target dataset and employs an additional adapter layer to enhance the representations of tail classes. VL-LTR~\cite{tian2022vl} presents a new visual linguistic framework for long-tailed visual recognition, which includes a class-wise text-image pre-training and a language-guided recognition head. However, they focus on leveraging the multi-modality capability of the vision-language framework, ignoring that long-tailed distribution can still corrupt the representation space.

\section{The proposed method}

\begin{figure*}[t]
  \centering
    \includegraphics[width=0.95\linewidth]{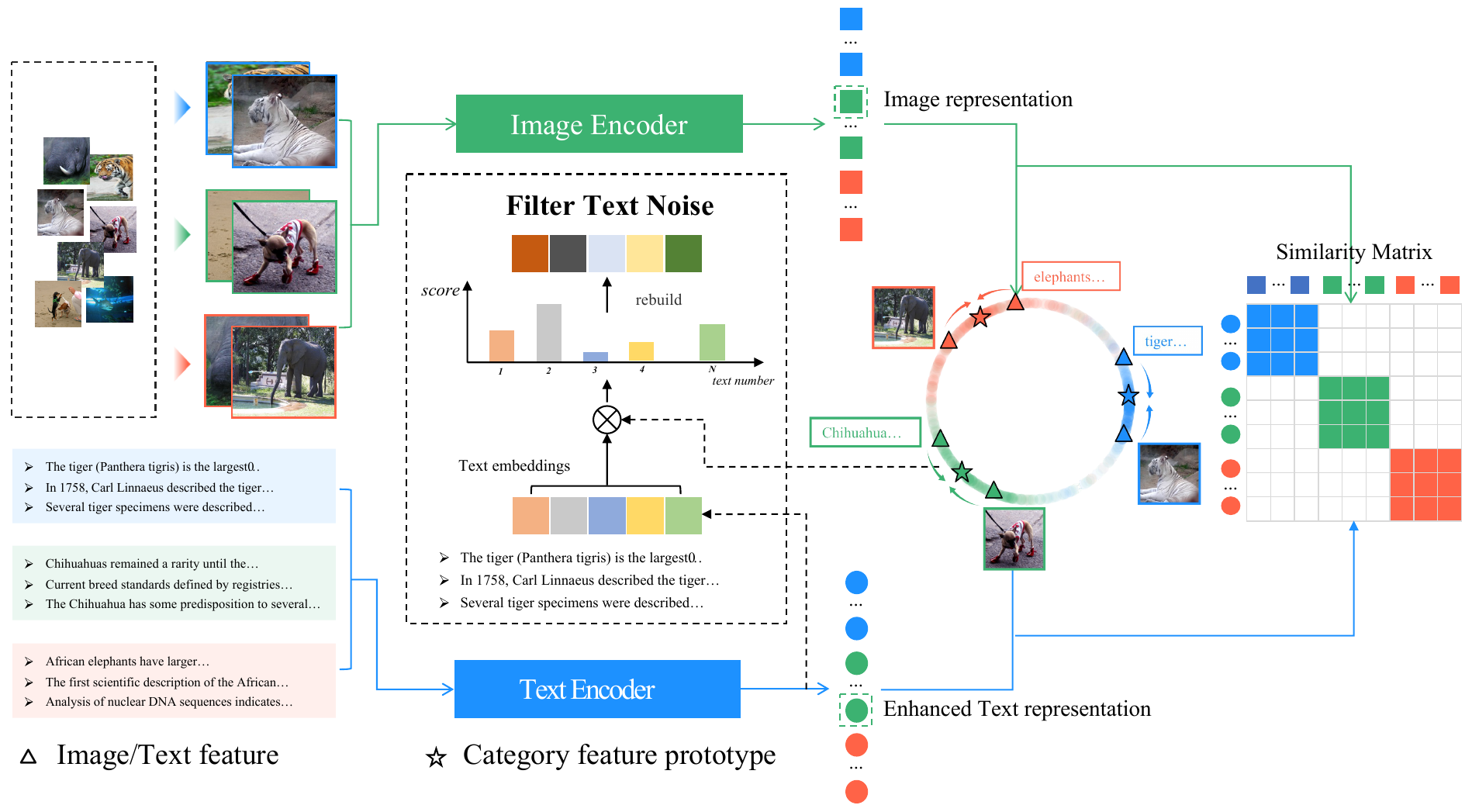}
    \caption{Overview of our proposed category-feature prototype guided image-text matching process.  Using a prototype-based contrastive loss function, we align the image and text features of the same category with category feature prototypes in the hypersphere space, which ensures an efficient representation of the feature space. To further enhance this process, we have also created a module to filter out irrelevant text and enhance attributes, mitigating noise interference in the text.  Together, these solutions mitigate issues in long-tailed recognition through balanced feature space and robust textual representations.}
    \label{figure2}
\end{figure*}

\subsection{Uniformly Distributed Category prototype Guided  Image-Text Matching}
In representation learning, an ideal characteristic is the attainment of prototypes with a uniform distribution~\cite{li2022targeted,cui2021parametric}. This property signifies that supervised contrastive learning should converge to prototypes with distinct classes that are uniformly dispersed on a hypersphere. A uniform prototype maximizes inter-class distances within the feature space, thereby optimizing the margin and enhancing the generalization capability of long-tailed recognition models. It has been observed that the attribute variability stemming from imbalanced data distribution in long-tailed recognition primarily contributes to reduced prediction confidence for tail classes. To tackle this issue, we devise an image and text matching mechanism founded upon the category prototype, which exhibits a uniform distribution on a hypersphere. The category prototype embodies the aforementioned class features. Our proposed method effectively mitigates feature space bias induced by imbalanced data distribution between head and tail classes, diverging from the traditional approach of directly comparing text and image information.

\par\noindent\textbf{Initialization of Uniformly Distributed Category Prototypes.}
In contrastive learning, ideal prototypes should be uniformly situated on the unit hypersphere $\mathbb{S}^{d-1}=\left\{\boldsymbol{u} \in \mathbb{R}^{d}:\|\boldsymbol{u}\|=1\right\}$ within the feature space. The robust representational capacity of the CLIP model enables the generation of uniformly distributed category features, which effectively represent distinct categories. Inspired by this, we employ the  CLIP model to pre-determine the optimal location for each category prototype in the hyper-spherical feature space. Specifically, we utilize the "anchor text" $T$, generated from a template in the format "a photo of a $\{label\}$", where $\{label\}$ is substituted with the category name, as the anchor text for each category. By feeding the anchor text into the pre-trained CLIP model, we obtain the "anchor text feature" and use it to initialize the category feature prototype as follows:

\begin{eqnarray}
\boldsymbol{c}^{k}=\mathcal{L}_{\mathrm{enc}}\left(T\right),
\label{equation1}
\end{eqnarray}

where $\boldsymbol{c}^{k}$ represents the prototype for category $k$, and $\mathcal{L}_{\mathrm{enc}}$ denotes the language encoder of the pre-trained CLIP model. We employ $\boldsymbol{c}^{k}$ as the target against which text and image information is matched, promoting the learning of more discriminative features by the text and image encoders. Furthermore, these initialized category feature prototypes are uniformly distributed on the hyper-sphere, preventing the encoders from developing a preference for high-frequency categories. It is worth noting that we use the pre-trained clip encoder in the process, and we will investigate its impact in Appendix.~\ref{appendixC} for more details.

\par\noindent\textbf{Uniformly Distributed Category Prototype-guided mechanism for image-text matching.}
Within vision-language frameworks, the prevalent method for learning feature representations involves directly aligning image and text features in the feature space. However, in long-tailed datasets, sample imbalances within tail categories may disrupt the uniformity of feature prototypes on the hyper-sphere space, resulting in performance deterioration. To tackle this challenge, we introduce an innovative approach based on category prototype contrastive learning to promote the alignment of image and text features. During the image-text matching process, a batch of images $\mathcal{I}=\{I_i\}_{i=1}^N$ and corresponding text sentences $\mathcal{T}=\{T_i\}_{i=1}^N$ are randomly sampled, where $N$ denotes the batch size. These images and texts are subsequently encoded into feature representations utilizing visual encoder $\mathcal{V}_{\mathrm{enc}}(\cdot)$ and language encoder $\mathcal{L}_{\mathrm{enc}}(\cdot)$, respectively:

\begin{eqnarray}
\boldsymbol{z}_{i}^{I}=\mathcal{V}_{\mathrm{enc}}\left(I_{i}\right), \quad \boldsymbol{z}_{i}^{T}=\mathcal{L}_{\mathrm{enc}}\left(T_{i}\right),
\label{equation2}
\end{eqnarray}

where the vectors $\boldsymbol{z}_{i}^{I}$ and $\boldsymbol{z}_{i}^{T}$ are both normalized. The category prototype contrastive loss function $\mathcal{L}_{\operatorname{PC}}$ can be defined as:

\begin{equation}
  \begin{aligned}
 \mathcal{L}_{\operatorname{PC}} =  - \log {\bigg[ \frac{\exp \left(\boldsymbol{z}_{i}^{I} \cdot \boldsymbol{c}^{i} / \tau\right)}{\sum_{j=1}^{K} \exp \left(\boldsymbol{z}_{i}^{I} \cdot \boldsymbol{c}^{j} / \tau\right)}\bigg]} 
  - \log {\bigg[ \frac{\exp \left(\boldsymbol{z}_{i}^{T} \cdot \boldsymbol{c}^{i} / \tau\right)}{\sum_{j=1}^{K} \exp \left(\boldsymbol{z}_{i}^{T} \cdot \boldsymbol{c}^{j} / \tau\right)}\bigg]},
 \end{aligned}
\end{equation}
 
where the parameter $\tau$, initialized to 0.07~\cite{khosla2020supervised}, is a hyper-parameter. The image-text matching process with category prototype contrastive loss encourages each class's image and text features to align with its corresponding uniformly distributed category prototype in the hypersphere. This alignment prevents the long-tailed distribution from breaking the uniformity of the feature space, thereby improving the model's representation quality of the long-tailed data. It encourages image and text features to move towards their corresponding category prototype while avoiding proximity to other category prototypes.

Moreover, the adaptation of category prototypes to the semantic and stylistic characteristics of the given dataset is crucial for enhancing the performance of the model. Thus, we propose a unique dynamic updating scheme for category feature prototypes that enables them to adjust to the imbalanced distribution of data in both the head and tail of the dataset. This method prevents the significant displacement of tail category feature prototypes, which would disrupt the uniformity of the prototypes. We implement an exponential moving average (EMA) algorithm~\cite{emanet} with class label frequency to update the category feature prototypes dynamically: 

\begin{equation}
  \begin{aligned}
    \boldsymbol{c}^{k} \leftarrow m \boldsymbol{c}^{k}&+(1-m) \bigg[\boldsymbol{z}_{i}^{T} + \pi_k \cdot \boldsymbol{z}_{i}^{I}\bigg]/(\pi_k + 1), \\ &\quad \forall i \in\left\{i \mid \hat{y}_{i}=k\right\},
    \label{equation4}
  \end{aligned}
\end{equation}

where $\pi$ denotes the sample frequency vector ($\pi_k=N_k/N$) and $m$ represents the momentum coefficient, which is typically set to 0.999. This algorithm ensures that the category feature prototypes remain stable and accurate, regardless of the variations in data distribution among different classes. The usage of class label frequencies allows for the prototypes to be updated more slowly with regard to the image information of the less frequently occurring classes, which helps prevent significant prototype deviations caused by scarce tail class image information.

\begin{figure*}[!t]
    \centering
    \includegraphics[width=\linewidth]{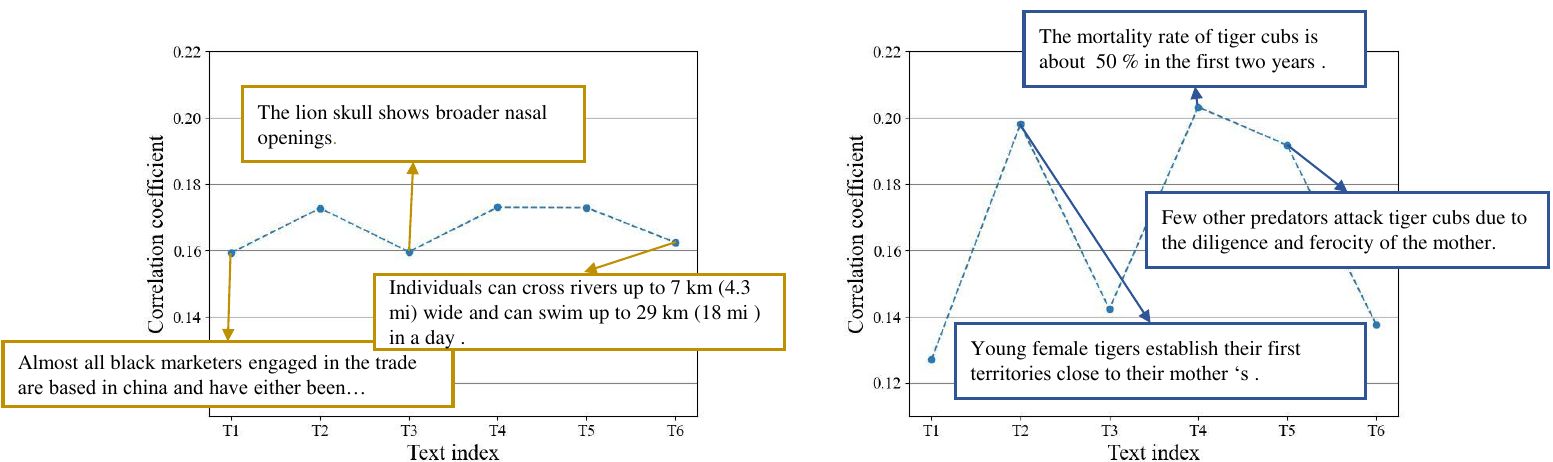}
    \caption{Comparison of text-image correlation coefficients (left) before and (right) after the application of the extraneous text filtering and attribute enhancement module for the tiger category in ImageNet-LT. It illustrates the text-image correlation coefficients for six paragraphs of text and a specific image of the corresponding category. The adjustment of weights for generic topics like "morphology," "source," and "root" highlights more relevant and specific attributes. }
    \label{figure3}
\end{figure*} 

\begin{figure*}[!t]
    \centering
    \subfigure{
    \centering \includegraphics[width=0.47\linewidth]{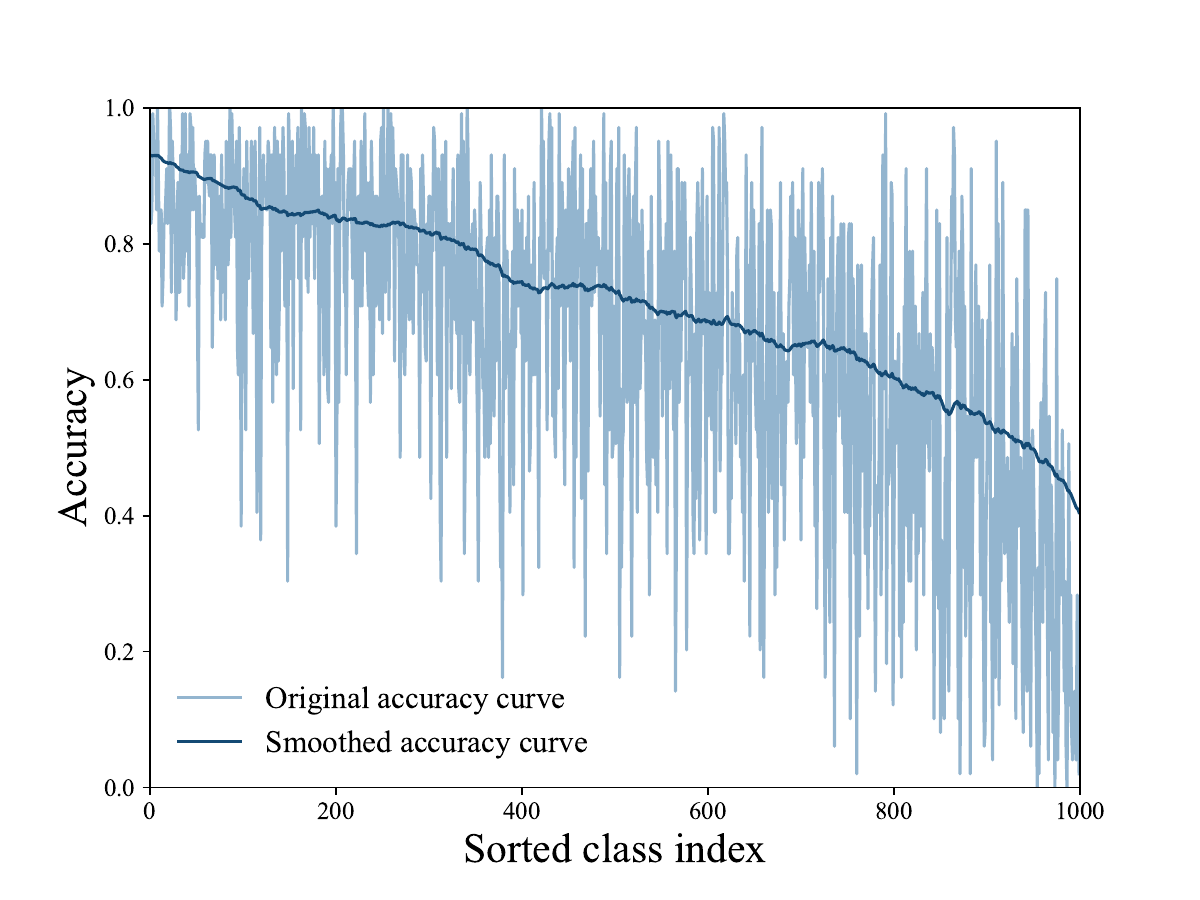}}
    \subfigure{
    \centering \includegraphics[width=0.47\linewidth]{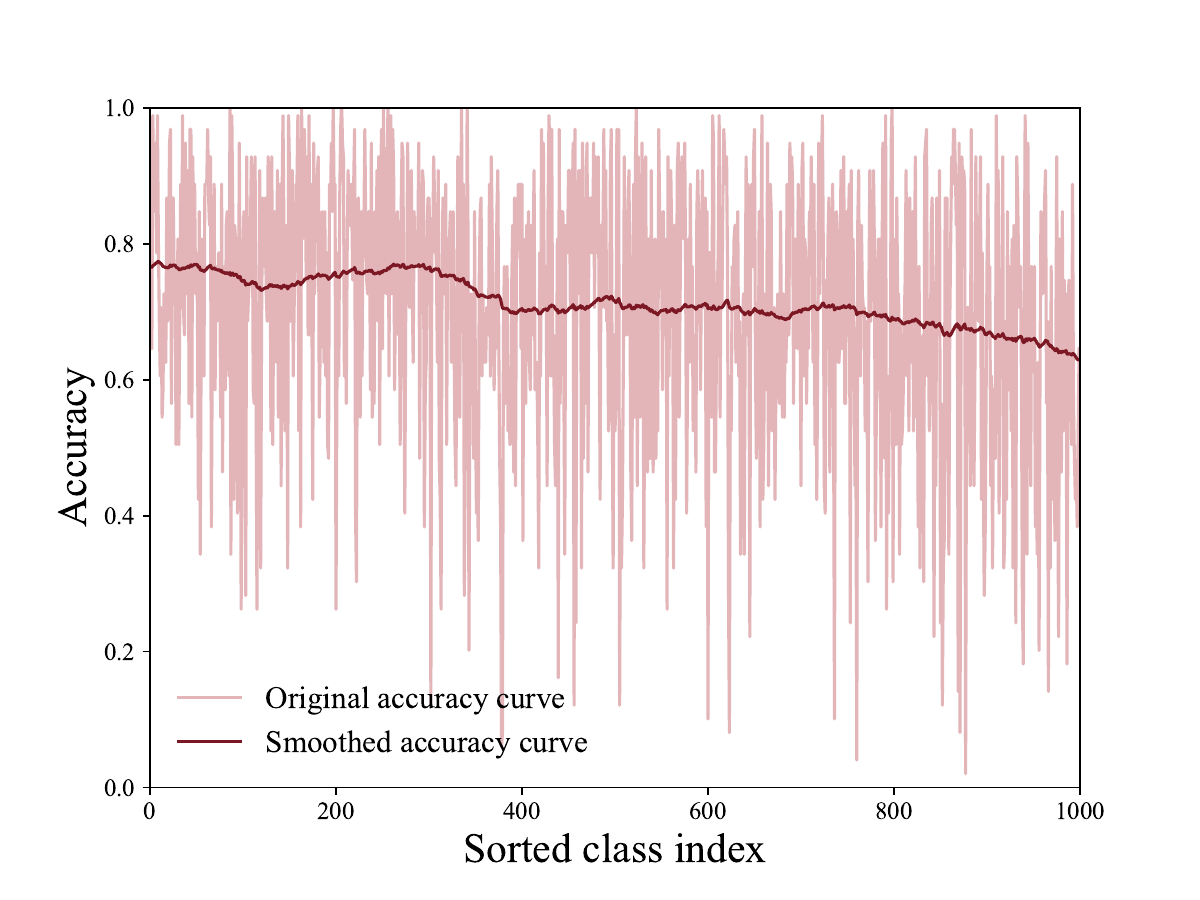}}
        \caption{ Comparison of learned classifier (left) and category prototype (right) accuracies with respect to the amount of class-specific training data available. Validation accuracy from a model trained on the ImageNet-LT. }
    \label{figure4}
\end{figure*} 

\par\noindent\textbf{The total loss function for image-text matching.}
During the image-text matching training phase, $\boldsymbol{z}_{i}^{I}$ and $\boldsymbol{z}_{j}^{T}$ are both d-dimension normalized vectors in the joint multimodal space. The overall loss function $\mathcal{L}_{\operatorname{Total}}$ is defined as:

\begin{equation}
  \begin{aligned}
     \mathcal{L}_{\operatorname{Total}} = & \mathcal{L}_{\text {ccl }}+\lambda \mathcal{L}_{\text {PC}}, \\
    \mathcal{L}_{\text {ccl }} = & -\frac{1}{\left|\mathcal{T}_{i}^{+}\right|} \sum_{T_{j} \in \mathcal{T}_{i}^{+}} \log \frac{\exp \left(\boldsymbol{z}_{i}^{I} \cdot \boldsymbol{z}_{j}^{T} / \tau\right)}{\sum_{T_{k} \in \mathcal{T}} \exp \left(\boldsymbol{z}_{i}^{I} \cdot \boldsymbol{z}_{k}^{T}  / \tau\right)} \\
    & -\frac{1}{\left|\mathcal{I}_{i}^{+}\right|} \sum_{I_{j} \in \mathcal{I}_{i}^{+}} \log \frac{\exp \left(\boldsymbol{z}_{i}^{T} \cdot \boldsymbol{z}_{j}^{I}  / \tau\right)}{\sum_{I_{k} \in \mathcal{I}} \exp \left(\boldsymbol{z}_{i}^{T} \cdot \boldsymbol{z}_{k}^{I}  / \tau\right)}, 
 \end{aligned}
\end{equation}

where $\mathcal{L}_{\text {ccl }}$ and $\mathcal{L}_{\text {PC }}$represent the conventional category-level contrastive learning loss function and the category feature prototype loss function we designed, respectively. $\lambda$ is the weight of the category feature prototype loss function. For detailed analysis on $\lambda$, please refer to the Appendix.~\ref{appendixD}. $\mathcal{T}_{i}^{+}$ represents a subset of $\mathcal{T}$ where each text shares the same category with the image $I_i$. Similarly, all images in $\mathcal{I}_i^+$ share the same category with the text $T_i$.

\setlength{\tabcolsep}{20pt}
\begin{table*}[!ht]
    \centering
    \caption{Results on ImageNet-LT in terms of accuracy (\%). $\dagger$ indicates the corresponding backbone is initialized with CLIP~\cite{radford2021learning} weights. $*$ represents the results of the reproduction of the method.}
        \begin{tabular}{llcccc}
            \toprule
            \multirow{2}{*}{Backbone} & \multirow{2}{*}{Method} & \multicolumn{4}{c}{Accuracy (\%)}                       \\ \cline{3-6} 
                                  &                         & Many & Med. & Few & All \\ \midrule
            ResNext-50 & SSD~\cite{li2021self}              & 66.8 & 53.1   & 35.4  & 56.0   \\
            ResNext-50                    & TADE~\cite{TADE}      & 66.5 & 57.0   &43.5 & 58.8   \\
            ResNext-50                    & RIDE (4 Experts)~\cite{RIDE}    & 68.2 & 53.8   & 36.0 & 56.8   \\
             ResNext-50                   &DisAlign~\cite{zhang2021distribution}                     &62.7  &52.1    &31.4  & 53.4   \\
            ResNext-101 & PaCo~\cite{cui2021parametric}                  &                     68.2 & 58.7   & 41.0 & 60.0 \\
            
            ResNext-101 &Res-LT~\cite{cui2022reslt}    &63.3 & 53.3   & 40.3 & 55.1  \\
            ResNext-152 & $\tau$-norm~\cite{2019Decoupling}         & 62.2 & 50.1   & 35.8 & 52.8 \\
            ResNext-152                  & NCM~\cite{2019Decoupling}                   & 60.3 & 49.0   & 33.6 & 51.3   \\
            ResNext-152                    &cRT~\cite{2019Decoupling} &64.7&49.1&29.4&52.4\\
            ResNext-152                    &LWS~\cite{2019Decoupling} &63.5&50.4&34.2&53.3\\\midrule
            ResNet-50$\dagger$  & $\tau$-norm~\cite{2019Decoupling}         & 60.9 & 48.4   & 33.8 & 51.2 \\
            ResNet-50$\dagger$                     & NCM~\cite{2019Decoupling}                   & 58.9 & 46.6   & 31.1 & 49.2  \\
            ResNet-50$\dagger$                     &cRT~\cite{2019Decoupling} &63.3&47.2&27.8&50.8\\
            ResNet-50$\dagger$                     &LWS~\cite{2019Decoupling} &62.2&48.6&31.8&51.5\\
            ResNet-50$\dagger$                     &vMF~\cite{wang2022towards} 
            &70.0 &	 58.4 &	 40.9 	& 60.5\\
            ResNet-50$\dagger$                    & Zero-Shot CLIP~\cite{radford2021learning}         & 60.8& 59.3   & 58.6 & 59.8  \\
            ResNet-50$\dagger$                     & Baseline               &74.4 & 56.9   &34.5 & 60.5   \\ 
            ResNet-50$\dagger$                     & BALLAD~\cite{ma2021simple}              &71.0 & 66.3 & 59.5 & 67.2   \\ 
            ResNet-50$\dagger$                      &VL-LTR~\cite{tian2022vl} &77.8 & 67.0 & 50.8 & 70.1  \\
            ResNet-50$\dagger$                      &VL-LTR$^\ast$~\cite{tian2022vl} &75.7 & 68.6 & 59.5 & 70.1  \\
            ResNet-50$\dagger$                          &\textbf{Ours}     & \textbf{76.4} & \textbf{69.5} & \textbf{60.2} & \textbf{70.9} \\
            \bottomrule
        \end{tabular}
    \label{table1}
\end{table*}

\subsection{Filtering Extraneous Text and Enhancing Attributes}
Owing to the noise present in internet text data, the extraction of core attribute semantics can be disrupted, leading to significant disparities between text and visual feature expressions. Consequently, we propose a text noise filtering module to augment our framework's robustness against noisy text data. Certain generic topics found in text descriptions, such as "morphology," "source," and "root," may be irrelevant to classification scenarios and could impact the modal alignment process within the feature space. To enhance and improve text feature representations, we introduce an extraneous text filtering and attribute enhancement module. Specifically, while constructing image-text pairs, we randomly sample $M$ description segments from the text corpus for an image of class $k$ and filter the candidate text using the category prototype feature. We reconstruct the text features by calculating the similarity score between the category prototype feature $\boldsymbol{c}_{k}$ and each text feature ${z}^{T, i}$:

\begin{eqnarray}
\begin{aligned}
z^T = \sum_{i=0}^M \frac{e^{\mathbf{z}^{T, i} \cdot \boldsymbol{c}^{k}}}{\sum_{j=1}^{M} e^{\mathbf{z}^{T, j} \cdot \boldsymbol{c}^{k}}}  \cdot \mathbf{z}^{T, i}.
\end{aligned}
\end{eqnarray}

The reconstruction of text feature representation via similarity score computation facilitates the model's focus on more critical information, thereby avoiding the learning of fixed and trivial correlations in specific image-text pairs. As illustrated in Fig.~\ref{figure3}, for the category (tail) of ImageNet-LT, the example demonstrates how reconstructing noisy text through similarity-based filtering enables the model to concentrate on the most vital information. Greater emphasis is placed on attribute details pertaining to taxonomy, ensuring that the model does not internalize a spurious correlation between the text and tiger images.

\setlength{\tabcolsep}{26pt}
\begin{table*}[!t]
    \centering
    \caption{Results on Place-LT in terms of accuracy (\%). $\dagger$ indicates the correspongding  backbone is initialized with CLIP~\cite{radford2021learning} weights. $*$ represents the results of the reproduction of the method.}
        \begin{tabular}{llcccc}
            \toprule
            \multirow{2}{*}{Backbone} & \multirow{2}{*}{Method} & \multicolumn{4}{c}{Accuracy (\%)}                       \\ \cline{3-6} 
                                  &                         & Many & Med. & Few & All \\ \midrule
            
            ResNext-152                    & TADE~\cite{TADE}       & 40.4   &43.2 & 36.8  & 40.9 \\
            
            ResNext-152 & PaCo~\cite{cui2021parametric}                  &                     36.1 & 47.9   & 35.3 & 41.2 \\
            
            ResNext-152 &Res-LT~\cite{cui2022reslt}    &39.8 & 43.6   & 31.4 & 39.8  \\
            ResNext-152 & $\tau$-norm~\cite{2019Decoupling}         & 37.8 & 40.7   & 31.8 & 37.9 \\
            ResNext-152                  & NCM~\cite{2019Decoupling}                   & 40.4 & 37.1   & 27.3 & 36.4   \\
            ResNext-152                    &cRT~\cite{2019Decoupling} &42.0&37.6&24.9&36.7\\
            ResNext-152                    &LWS~\cite{2019Decoupling} &40.6&39.1&28.6&37.6\\\midrule
            ResNet-50$\dagger$  & $\tau$-norm~\cite{2019Decoupling}         & 34.5 & 31.4   & 23.6 & 31.0 \\
            ResNet-50$\dagger$                     & NCM~\cite{2019Decoupling}                   & 37.1 & 30.6   & 19.9 & 30.8  \\
            ResNet-50$\dagger$                     &cRT~\cite{2019Decoupling} &38.5&29.7&17.6&30.5\\
            ResNet-50$\dagger$                     &LWS~\cite{2019Decoupling} &36.0&32.1&20.7&31.3\\
                                
             ResNet-50$\dagger$                    & Zero-Shot CLIP~\cite{radford2021learning}         & 37.5& 37.5   & 40.1 & 38.0  \\
             ResNet-50$\dagger$                     & Baseline               &50.8 & 38.6   &22.7 & 39.7   \\ 
             ResNet-50$\dagger$                     & BALLAD               &46.7 & 48.0 & 42.7 & 46.5   \\ 
            ResNet-50$\dagger$                      &VL-LTR~\cite{tian2022vl} &51.9 & 47.2 & 38.4 & 48.0  \\
            ResNet-50$\dagger$                      &VL-LTR$^\ast$~\cite{tian2022vl} &50.0 & 48.3 & 44.2 & 48.0  \\
            ResNet-50$\dagger$                          &\textbf{Ours}     & \textbf{50.8} & \textbf{48.8} & \textbf{44.6} & \textbf{48.7} \\
            \bottomrule
        \end{tabular}
    \label{table2}
\end{table*}

\subsection{Enhancing Image Recognition through Category Prototype-Guided Recognition}

In the image recognition stage, taking the linear classifier as an example, the weight of the classifier is defined as $\omega_i, \{i = 0,\ldots, K\}$, and the probability of a sample belonging to the $i$-th class is given by:

\begin{equation}
    P^l_i = \frac{\exp(\omega_{i} \cdot \boldsymbol{z}^I + b_i)}{\sum_{j=0}^{K} \exp(\omega_{j} \cdot \boldsymbol{z}^I + b_j)}
\end{equation}

 However, due to the long-tail data distribution, as shown in Fig.~\ref{figure4} (left), the classification accuracy of the linear classifier for each class (the number of samples for each class is arranged in descending order on the x-axis) is significantly higher for the head classes than the tail classes, known as the forward bias of the classifier towards the head classes~\cite{wang2022towards}.

The classifier based on class feature prototypes uses the distance between the sample features and category prototypes to classify the samples. The formula for classifying samples can be expressed as follows:

\begin{equation}
    P_i^p = \frac{\exp \left(\boldsymbol{z}^{I} \cdot \boldsymbol{c}^{i} \right)}{\sum_{j=1}^{K} \exp \left(\boldsymbol{z}^{I} \cdot \boldsymbol{c}^{j} \right)}
\end{equation}

Fig.~\ref{figure4} (right) shows the classification results based on the category prototypes (with the number of samples per class on the x-axis), compared to Fig.~\ref{figure4} (left), the performance difference between classes is smaller, but the positive bias towards the head classes is reduced. Inspired by this phenomenon, we propose a classifier structure based on guiding the class feature prototypes. Specifically, we combine the pre-trained feature category prototypes $\boldsymbol{c}^{k}$ on the long-tailed dataset with the learnable classifier. The probability of a sample belonging to class $i$ is calculated as follows:

\begin{equation}
    P_i = \alpha \cdot P_i^p + (1 - \alpha) \cdot P_i^l
\end{equation}

where $P_i^p$ is the output result of the prototype classifier, and $P_i^l$ is the output result of the learnable classifier. Our designed classifier output result includes both the positive bias towards head classes and the performance compensation for tail classes. We will investigate the impact of $\alpha$ in Appendix.~\ref{appendixE} for more details. 

\section{Experiments}
In this section, a series of experiments were conducted to validate the effectiveness of the proposed method in image classification tasks, including an analysis of experiments and ablation studies.

\subsection{Datasets and setup}
Three long-tailed recognition benchmark datasets were used in this study, including ImageNet-LT~\cite{openlongtailrecognition}, Places-LT~\cite{zhou2017places}, and iNaturalist 2018~\cite{van2018inaturalist}. More details about datasets are shown in Appendix.~\ref{appendixA}.

\subsection{Experimental Results on ImageNet-LT}
\noindent\textbf{Experimental Setup.}
\label{Setup}
To validate the effectiveness of our proposed method, we conducted extensive experiments on ImageNet-LT~\cite{openlongtailrecognition}. Following the settings of previous work~\cite{tang2020long}~\cite{LADE}, we employed ResNet-50~\cite{resnet} as the visual encoder and a 12-layer Transformer~\cite{radford2019language} as the language encoder. More details are shown in Appendix.~\ref{appendixA}.

To make a fair comparison, we also built a baseline method that only used visual modality, which had the same settings as our proposed method except that the baseline model was directly initialized with CLIP pre-trained weights and fine-tuned for 100 epochs. In addition, we also reproduced and reported the performance of some representative methods such as $\tau$-normalized, cRT, NCM, and LWS \cite{2019Decoupling}, which were all initialized with CLIP pre-trained weights.

\par\noindent\textbf{Performance comparison on the ImageNet-LT validation set.}
This section presents the performance comparison of our proposed method on the ImageNet-LT dataset. To ensure a fair comparison, we used various methods from the deep long-tailed learning survey~\cite{longtailedsurvey}. Tab.~\ref{table1} shows the results of our model, which outperforms traditional visual methods that use similar visual encoders (i.e., backbone networks). For instance, when using ResNet-50 \cite{resnet} as the backbone network, our method achieved an overall accuracy of 70.9\%, which is 10.4 percentage points higher than the baseline method (70.9\% vs. 60.5\%). Furthermore, our method outperformed the previously best-performing methods VL-LTR \cite{tian2022vl} and BALLAD~\cite{ma2021simple} by 0.8 percentage points (70.9\% vs. 70.1\%) and 3.7 percentage points (70.9\% vs. 67.2\%), respectively. In addition, from the perspective of tail category accuracy, our method also performed well, achieving a 0.7 percentage point improvement over the second-best method \cite{tian2022vl} (60.2\% vs. 59.5\%).

\subsection{Results on Place-LT Dataset}
\par\noindent\textbf{Performance comparison on the Place-LT validation set.} Tab.~\ref{table2} presents results on the Place-LT dataset using ResNet-50. Our model achieved an overall accuracy of 48.7\%, surpassing other state-of-the-art methods by at least 7.5 percentage points (48.7\% compared to 41.2\%), including PaCo \cite{cui2021parametric}, TADE \cite{TADE}, and ResLT \cite{cui2022reslt}, which all use ResNet-152 \cite{resnet} as the backbone network. Compared to other CLIP-based methods such as BALLAD \cite{ma2021simple} and VL-LTR \cite{tian2022vl}, our proposed method still exhibited competitive performance (48.7\% compared to 46.5\% and 48.0\%, respectively).

\setlength{\tabcolsep}{10pt}
    \begin{table}[t]
    \centering
    \caption{Results on iNaturalist 2018 in terms of accuracy (\%). $\dagger$ indicates the  backbone is initialized with CLIP~\cite{radford2021learning} weights.}
        \begin{tabular}{llc}
        \toprule
        Backbone& Method & Accuracy (\%)                       \\  \midrule
        ResNext-50 & SSD~\cite{li2021self}              & 71.5   \\
        ResNext-50                    & TADE~\cite{TADE}      & 72.9\\
        ResNext-50                    & RIDE (4 Experts)~\cite{RIDE}    & 72.6   \\
        ResNext-50 & PaCo~\cite{cui2021parametric}                  &                   73.2  \\
        
        ResNext-50 &Res-LT~\cite{cui2022reslt}    &72.3 \\
        ResNext-50 & $\tau$-norm~\cite{2019Decoupling}         & 69.3  \\
        ResNext-50                  & NCM~\cite{2019Decoupling}                   & 63.1  \\
        ResNext-50                   &cRT~\cite{2019Decoupling} &67.6\\
        ResNext-50                    &LWS~\cite{2019Decoupling} &69.5\\\midrule
        ResNet-50$\dagger$  & $\tau$-norm~\cite{2019Decoupling}         & 71.2\\
        ResNet-50$\dagger$                     & NCM~\cite{2019Decoupling}                   & 65.3 \\
        ResNet-50$\dagger$                     &cRT~\cite{2019Decoupling} &69.9\\
        ResNet-50$\dagger$                     &LWS~\cite{2019Decoupling} &71.0\\
                            
         ResNet-50$\dagger$                    & Zero-Shot CLIP~\cite{radford2021learning}         & 3.4  \\
        ResNet-50$\dagger$                     & Baseline               &72.6  \\ 
        ResNet-50$\dagger$                      &VL-LTR~\cite{tian2022vl} &74.6 \\
        ResNet-50$\dagger$                          &\textbf{Ours}     & \textbf{75.2}  \\
                             \bottomrule
        \end{tabular}
    \label{table3}
\end{table}

\subsection{iNaturalist 2018 Experiment Results}
\par\noindent\textbf{Performance Comparison on the iNaturalist 2018 Validation Set.} 
Tab.~\ref{table3} shows the top-1 accuracy of different methods on the iNaturalist 2018 dataset. We found that when using ResNet-50~\cite{resnet} as the backbone network, our model achieved an overall accuracy of 75.2\%, surpassing previous methods with the same backbone network by at least 0.6 percentage points.

\subsection{Ablation studies} 
We conducted ablation studies on ImageNet-LT to further clarify the choice of hyperparameters in our method and the effect of each component. All experiments were conducted under the same settings as Section ~\ref{Setup} for a fair comparison. In addition, we provide the feature map t-SNE visualization about our method in Appendix.~\ref{appendixF}.

\par\noindent\textbf{Impacts of category-feature prototype initialization strategies.}
We studied the effectiveness of the “anchor text” selection strategy by replacing each category's "a photo of a \{label\}" text form with a "cut-off" strategy. In this strategy, we simply select the first $six$ sentences from the text description as the anchor text input to the pre-trained CLIP model to generate category features and then cluster to form category feature prototypes.  As shown in the first and second rows of Tab.~\ref{table4}, the model using the "a photo of a \{label\}" text form achieved 1.4\% higher overall accuracy (70.9\% vs. 69.5\%) than the model using the "cut-off" text form. This demonstrates the effectiveness of the "a photo of a \{label\}" text form selection in initializing category feature prototypes.

\par\noindent\textbf{Effectiveness of extraneous text filter module.}
The effectiveness of the unrelated text filtering module was verified by comparing it with a model that did not use the module. As shown in Tab.~\ref{table5} rows 1 and 3, the model without the unrelated text filtering module experienced a 1.1\% decrease in overall accuracy compared to the model with the module (70.9\% vs. 69.8\%). This demonstrates that the unrelated text filtering module can effectively reconstruct the text information and improve the model's attention to important attributes in the text information.

\par\noindent\textbf{Effectiveness of the category prototype contrastive loss function.}
We also tested the effectiveness of the contrastive loss function for the category feature prototype by removing it during the text-image feature alignment stage and using a regular category-level contrastive loss function instead. As shown in Tab.~\ref{table5} rows 1 and 2, the category feature prototype contrastive loss function significantly improved the model's performance, by approximately 3.8\% (70.9\% vs. 67.1\%). This indicates that the category feature prototype contrastive loss function can effectively ensure the uniformity of the model's feature space and prevent the degradation of feature representation quality due to the long-tailed data distribution.

\setlength{\tabcolsep}{13pt}
\begin{table}[t]
  \centering
  \caption{Ablation experiments on the effectiveness of Prototypes Initialization strategy and  extraneous text filter module under ImageNet-LT.}
        \begin{tabular}{ccc}
            \toprule
            Prototypes Initialization & Text Filter & Accuracy \\
            \midrule
            label & $\checkmark$ & 70.9 \\
            cut off & $\checkmark$ & 69.5 \\
            label & - & 69.8 \\
            \bottomrule
        \end{tabular}
        \label{table4}
\end{table}

\setlength{\tabcolsep}{6pt}
\begin{table}[t]
  \centering
  \caption{Ablation experiments on the effectiveness of each component in our method under ImageNet-LT.  "Pro Cl" represents the category prototype-guided recognition classifier module in the image recognition stage.}
    \begin{tabular}{ccccc}
        \toprule
        \multicolumn{2}{c}{Text-image Matching} &\multicolumn{2}{c}{Image Recognition}  & \multirow{2}{*}{Accuracy} \\
        \cmidrule(lr){0-1} \cmidrule(lr){3-4}
          w/o $\mathcal{L}_{\text {PC}}$  & w/ $\mathcal{L}_{\text {PC}}$  &  w/o Pro Cl & w/ Pro Cl & \\
          \midrule
          -  & $\checkmark$   & - & $\checkmark$ & 70.9\\
          $\checkmark$ & -     & - & $\checkmark$ & 67.1\\
          - & $\checkmark$     & $\checkmark$  & - & 65.9\\
        \midrule
    \end{tabular}
    \label{table5}
\end{table}

\par\noindent\textbf{Effectiveness of the category feature prototype-guided classifier design module.} Finally, we tested the effectiveness of the category feature prototype-guided classifier by comparing it with a standalone category feature prototype classifier and a linear classifier. As shown in Tab.~\ref{table5} rows 1 and 3, the proposed category feature prototype-guided classifier improved the overall accuracy by 5.0\% compared to the standalone linear classifier (70.9\% vs. 65.9\%). These results demonstrate that the category feature prototype-guided classifier design module can effectively prevent positive bias in the classifier and highlight the effectiveness of the module.

\section{Conclusion}
In this paper, we propose a novel visual-linguistic representation framework based on the category prototype guidance to alleviate the long-tail problem.  In the text-image modality matching training stage, we design a category prototype contrastive loss function to align text and image features onto category feature prototypes uniformly distributed in feature space. Additionally, our proposed irrelevant text filtering and attribute enhancement module allows the model to ignore the irrelevant noisy text and focus more on key attribute information, thereby enhancing the robustness of our framework. In the image recognition fine-tuning stage, we design a classifier structure based on the category feature prototype, which compensates for the performance of tail classes while maintaining the performance of head classes. Our proposed method achieves state-of-the-art performance on long-tailed distributed datasets, including ImageNet-LT, Places-LT, and iNaturalist 2018.

\section*{Acknowledgements}
We want to extend our heartfelt thanks to everyone who has provided their suggestions regarding our manuscript. Their insightful comments and constructive feedback have significantly contributed to the improvement of this paper.

\bibliographystyle{ACM-Reference-Format}
\balance
\bibliography{sample-base}

\newpage
\appendix

\section{Datasets and setup. } 
\label{appendixA}
\par\noindent\textbf{DATASETS. } Three challenging long-tailed visual recognition benchmark datasets were used in this study, including ImageNet-LT~\cite{openlongtailrecognition}, Places-LT~\cite{zhou2017places}, and iNaturalist 2018~\cite{van2018inaturalist}. ImageNet-LT, which consists of 1000 categories, was constructed by sampling a subset of ImageNet-2012 according to the Pareto distribution (power value $\alpha=6$). The training set contains 115.8K images, with the number of images per category ranging from 1280 to 5. The validation set and test set are balanced, containing 20K and 50K images, respectively. Similar to ImageNet-LT, Places-LT is a long-tailed version of the large-scale scene classification dataset Places, with 365 categories and 62.5K images, with the cardinality of categories ranging from 5 to 4980. iNaturalist 2018 is a real-world natural long-tailed dataset consisting of 8142 fine-grained species. The training set contains 437.5K images, with an imbalance factor of 500. The official validation set containing three images per category was used to test the proposed method. The validation set was used to select hyperparameters, and the test set was used to report numerical results.

Following the setting of VL-TLR~\cite{tian2022vl}, we also collected category-level textual descriptions for these three datasets. The textual descriptions were mainly obtained from Wikipedia, an open-source online encyclopedia containing millions of free articles. We first used the original class names as initial queries to obtain the best-matching entries on Wikipedia. After cleaning and filtering out some obviously irrelevant parts of these entries, such as "References" or "External Links," we split the remaining parts into sentences to form the original text candidate set for each category. It should be noted that some categories had relatively few sentences, so we added 80 additional prompt sentences for each category to alleviate the problem of data scarcity. These prompt sentences were generated using the form of "a photo of a {label}" based on the prompt template provided in prior works~\cite{radford2021learning}.

\par\noindent\textbf{Experimental Setup. }To validate the effectiveness of our proposed method, we conducted extensive experiments on ImageNet-LT~\cite{openlongtailrecognition} using PyTorch. Following the settings of previous works~\cite{tang2020long}~\cite{LADE}, we employed ResNet-50~\cite{resnet} as the visual encoder and a 12-layer Transformer~\cite{radford2019language} as the language encoder. 

All models were optimized using AdamW optimizer with a momentum of 0.9 and weight decay of $5 \times 10^{-2}$. We used the same data augmentation method as previous works~\cite{touvron2021training}. In the pre-training stage, the maximum length of text tokens was set to 77 (including [SOS] and [EOS] tokens), and we loaded the pre-trained weights of CLIP~\cite{radford2021learning}. The initial learning rate was set to $5 \times 10^{-5}$ and followed the cosine annealing~\cite{sgdr}. The model was trained for 50 epochs in the first stage of cross-modal feature alignment with a batch size of 256. In the second-stage image recognition, we set the batch size to 128 and fine-tuned the model for another 50 epochs. We set the initial learning rate to $1 \times 10^{-3}$ and still used cosine annealing for decay. Unless otherwise specified, we used an input size of 224 × 224 and the square-root sampling strategy~\cite{mahajan2018exploring}~\cite{mikolov2013distributed} in both stages. 

We also evaluated our proposed method on the Place-LT dataset, which includes images from different domains~\cite{zhou2017places}. The experimental setup for the Place-LT dataset is the same as experiments in ImageNet-LT.

We further tested our approach on the iNaturalist 2018. Following the conventional practice \cite{touvron2021training}, we trained our model for 100 epochs in the first stage and then 360 epochs for image classification in the second stage. The initial learning rates for the pre-training and fine-tuning stages were set to $5 \times 10^{-4}$ and $2 \times 10^{-5}$, respectively. The same training stages and settings were used for the baseline model as those proposed in our approach. All other experimental settings were the same as those in ImageNet-LT.

\section{The qualitative experiment analysis of the toy feature distribution. }
\label{appendixB}
To further demonstrate the qualitative experiment analysis of the toy feature distribution shown in Fig.~\ref{figure1}. In Tab.~\ref{table6}, we compare the alignment and neighborhood uniformity,  of the method before and after implementing the uniformly distributed category prototype-guided image-text matching strategy on ImageNet-LT. The method without implementing the uniformly distributed category prototype-guided image-text matching strategy is named baseline (the same as the "Baseline" in Tab.~\ref{table1}).  Similar to \cite{wang2020understanding}, we define alignment under our setting as the average distances between samples from the same class, where $F_i$ is the set of features from class $i$:

\begin{equation}
    \mathbf{A}=\frac{1}{C} \sum_{i=1}^{C} \frac{1}{\left|F_{i}\right|^{2}} \sum_{v_{j}, v_{k} \in F_{i}}\left\|v_{j}-v_{k}\right\|_{2}.
\end{equation}

We define neighborhood uniformity as the distance to the top-$k$ closest class centers of each class:

\begin{equation}
\mathbf{U}_{k}=\frac{1}{C k} \sum_{i=1}^{C} \min _{j_{1}, \cdots, j_{k}}\left(\sum_{l=1}^{k}\left\|c_{i}-c_{j_{l}}\right\|_{2}\right),
\end{equation}

where $j_1, \ldots , j_k$ represent different classes. And $c_i$ is the center of samples from class $i$ on the hypersphere: $c_i = \frac{\sum_{v_j \in F_i} v_j}{||\sum_{v_j \in F_i} v_j||^2}$. The results underscore several advantageous properties of our method:

1.) Our method achieves superior neighborhood uniformity across all class splits and maintains better alignment with the baseline.

2.) The neighborhood uniformity (the average uniformity of the closest 10 classes) is 0.09 higher than that of the baseline. Furthermore, the baseline is even worse on tail classes, while our method maintains consistent neighborhood uniformity across all classes. This demonstrates the effectiveness of our method in keeping all classes separate from each other, thereby allowing for clearer decision boundaries between classes.

These results suggest that our method is more effective for handling the long-tailed distribution of the ImageNet-LT dataset. Our method may provide superior image-text matching performance under such circumstances, maintaining consistency and clear decision boundaries among all classes.

\setlength{\tabcolsep}{8pt}
\begin{table}[t]
    \begin{center}
    \caption{Implementing the uniformly distributed category prototype-guided image-text matching strategy achieves better alignment and neighborhood uniformity than the baseline on ImageNet-LT. The $k$ for neighborhood uniformity is set to 10. $\uparrow$ indicates larger is better, whereas
    $\downarrow$ indicates smaller is better.
      }
        \begin{tabular}{l|l|cccc}
            \toprule
            \multirow{2}{*}{Metric} &\multirow{2}{*}{Methods} & \multicolumn{4}{c}{ImageNet-LT}   \\ \cmidrule{3-6}  & & Many & Med. & Few &All  \\ \midrule
              \multirow{2}{*}{$\mathrm{A}$ $\downarrow$} &Baseline              &  0.64& 0.63 & 0.67&0.64   \\
              
               &Ours &0.59& 0.61&0.62&0.60\\
            
                              \midrule
              \multirow{2}{*}{$\mathrm{U}_{10}$ $\uparrow$} &Baseline              &  0.95& 0.93  & 0.92&0.94   \\
              
               &Ours &1.04&1.03&1.01&1.03\\
            \midrule
                               \multirow{2}{*}{$\mathrm{Acc.}$ $\uparrow$} &Baseline               &74.4 & 56.9   &34.5 & 60.5 \\
              
               &Ours & \textbf{76.4} & \textbf{69.5} & \textbf{60.2} & \textbf{70.9}\\
                                 \bottomrule
        \end{tabular}
    \label{table6}
    \end{center}
\end{table}
\setlength{\tabcolsep}{10pt}
\begin{table}[t]
  \centering
  \caption{Ablation experiments on the effectiveness of Prototypes Initialization strategy with the pre-trained clip weight under ImageNet-LT.}
        \begin{tabular}{ccc}
            \toprule
            Prototypes Initialization & weight & Accuracy \\
            \midrule
            label & pre-trained CLIP & 70.9 \\
           
            label & randomly initial  & 63.8 \\
            \bottomrule
        \end{tabular}
        \label{table7}
\end{table}

\section{Impact of the pre-trained clip weight for initialization of uniformly distributed category prototypes. }
\label{appendixC}

To investigate the effect of leveraging CLIP pre-trained weights in the initialization of uniformly distributed category prototypes, we undertake an experiment where we initialize the category prototypes using randomly initialized weights. A comparison between entries $\#2$ and $\#3$ in Tab.~\ref{table7} indicates that initializing the category prototypes with CLIP pre-trained weights confers a clear advantage. In addition, the large margin proves that the CLIP pre-trained weights generate the uniformly distributed category prototype, but the randomly initialized weights generate the biased  category prototype, leading to performance degradation.

\section{Ablation study of \texorpdfstring{$\lambda$}{} in Eq. 5. }
\label{appendixD}
 Moreover, in Eq.~5, we introduce a scaling hyperparameter $\lambda$, which governs the degree of the $\mathcal{L}_{\text {PC}}$. Initially, $\lambda$ is set to 0.1 and we vary the hyperparameter $\lambda$ within the range of 0.1 to 0.9, using a stride of 0.2, and perform ablation experiments with the resulting five sets of values, as shown in Fig.~\ref{figure5}. In general, a larger $\lambda$ reflects a higher level of confidence in adjusting the text-image matching process. The optimal $\lambda$ for ImageNet-LT is established at 0.5.

\section{Ablation study of \texorpdfstring{$\alpha$}{} in Eq. 9. }
\label{appendixE}
\begin{figure}[!t]
\centering
\includegraphics[width=\linewidth]{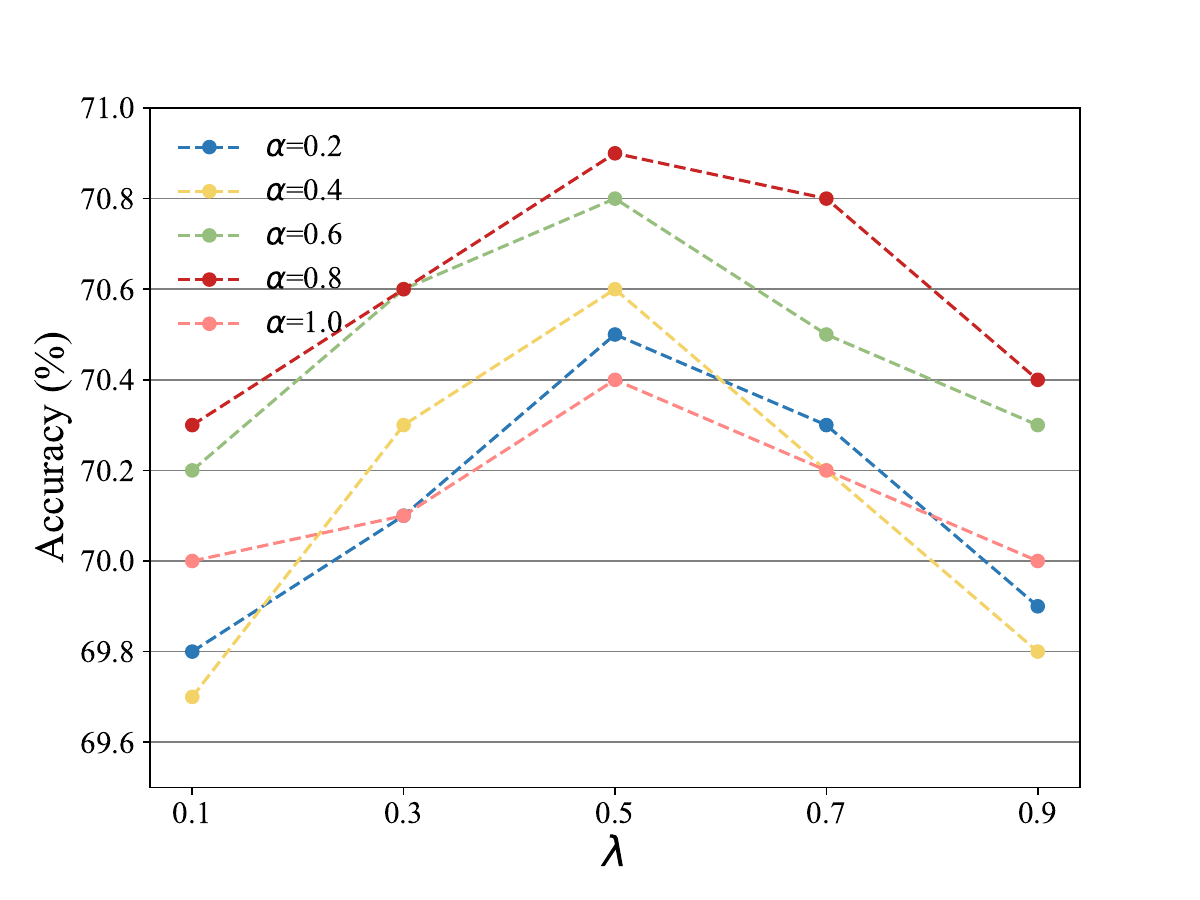}
\caption{Ablation study on hyper-parameters.}
    \label{figure5}
\end{figure}

Our method involves a crucial hyper-parameter, $\alpha$, in Eq.~9, which serves to balance the contributions of the two classification prediction terms $P_i^p$ and $P_i^l$. We evaluate the hyper-parameter $\alpha$ across the following values: $\alpha \in \left\{0.2, 0.4, 0.6, 0.8, 1.0\right\}$. We investigate the sensitivity of the model's accuracy to these different $\alpha$ values. Fig.~\ref{figure5} demonstrates the influence of the trade-off parameter $\alpha$ on accuracy. The results indicate that optimally combining $P_i^p$ and $P_i^l$ with $\alpha$ set at 0.8 yields the best performance.

\section{T-SNE visualization and the cross modalities similarity map analysis. }
\label{appendixF}
\begin{figure}[h]
\centering
\subfigure{
\centering \includegraphics[width=0.48\linewidth]{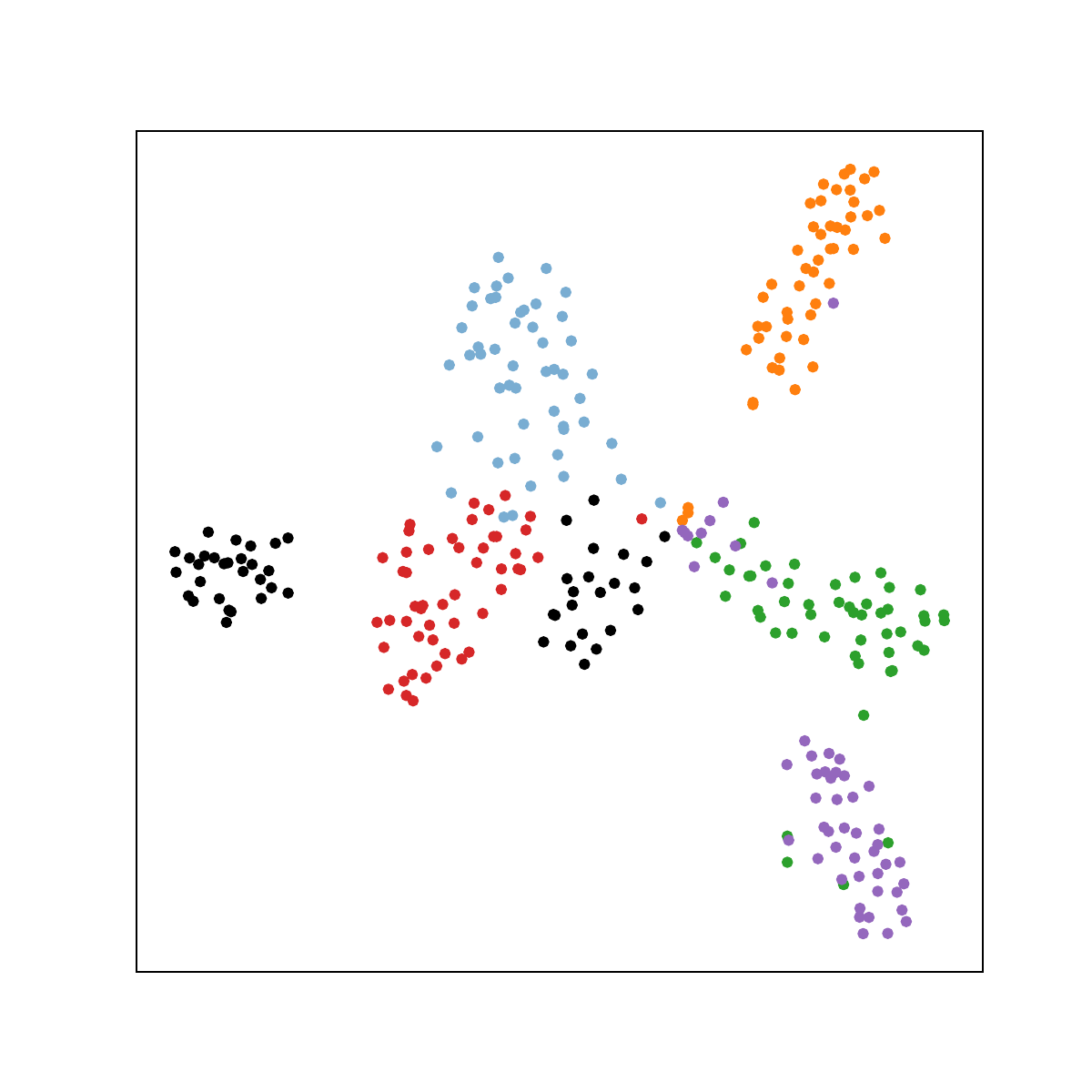}}
\subfigure{
\centering \includegraphics[width=0.48\linewidth]{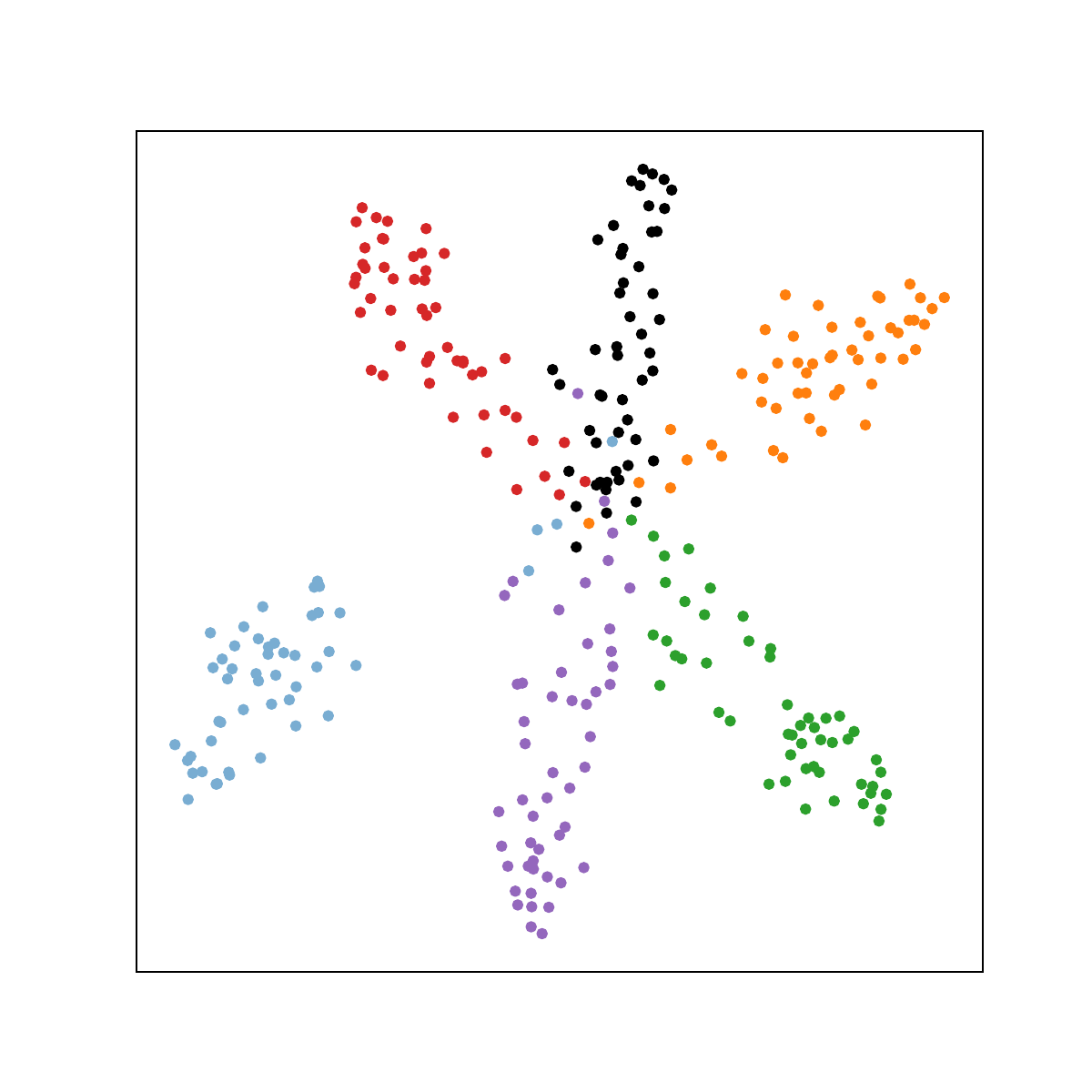}}
    \caption{t-SNE visualization comparion (left) VL-TLR~\cite{tian2022vl} and (right) ours on ImageNet-LT. We randomly extract fifty features from six categories in the test set. 
    }
    \label{figure6}
\end{figure}

To further demonstrate the effectiveness of our methods, we show the T-SNE visualization as Fig.~\ref{figure6}. As shown in Fig.~\ref{figure6}, the representation space of VL-TLR is dominated by the head category, leading to poor uniformity in learned category prototypes. Our approach improves performance by enhancing uniformity. 

\end{document}